%% file: 00_main.tex
\documentclass[letterpaper, 10 pt, conference]{ieeeconf}

\usepackage{breqn}
\usepackage{cuted}
\usepackage{xcolor}
\usepackage{capt-of}
\usepackage{amsfonts}
\usepackage{hyperref}
\usepackage{multirow}
\usepackage{mathtools}
\usepackage{dblfloatfix}
\usepackage[table]{xcolor}
\usepackage{booktabs}
\usepackage{makecell}
\usepackage{amsmath}
\usepackage{overpic}
\usepackage{amsfonts}
\usepackage{amssymb}
\usepackage{graphicx}
\usepackage{algorithm}
\usepackage{hyperref}
\usepackage{amsmath}
\usepackage{algpseudocode}
\usepackage{booktabs}
\usepackage[normalem]{ulem}
\usepackage[font=small, labelfont=bf]{caption}

\usepackage{amssymb}

\usepackage{fancyhdr} 
\renewcommand{\arraystretch}{1.2}

\definecolor{colorTrd}{rgb}{0.95,0.95,0.65}
\definecolor{colorSnd}{rgb}{1, 0.85, 0.7}

\definecolor{pink}{RGB}{255, 192, 203}

\newcommand{\del}[1]{}

\newcommand{\best}[1]{\cellcolor{colorTrd}\textbf{#1}}

\usepackage[T1]{fontenc}      
\usepackage[utf8]{inputenc}   
\usepackage[vietnamese]{babel}

\title{\LARGE \bf Geometry-Aware Online Mapping for 3D Gaussian~Splatting~SLAM}

\author{
    Thai Luu$^{1}$%
    \and Quan Tran$^{2}$%
    \and Hieu Phan$^{1}$%
    \and Tuan Dang$^{*1}$%
}

\begin{document}


\twocolumn[{%
    \renewcommand\twocolumn[1][]{#1}

    \maketitle

    \begin{center}
        \centering
        \begin{overpic}[width=\linewidth]{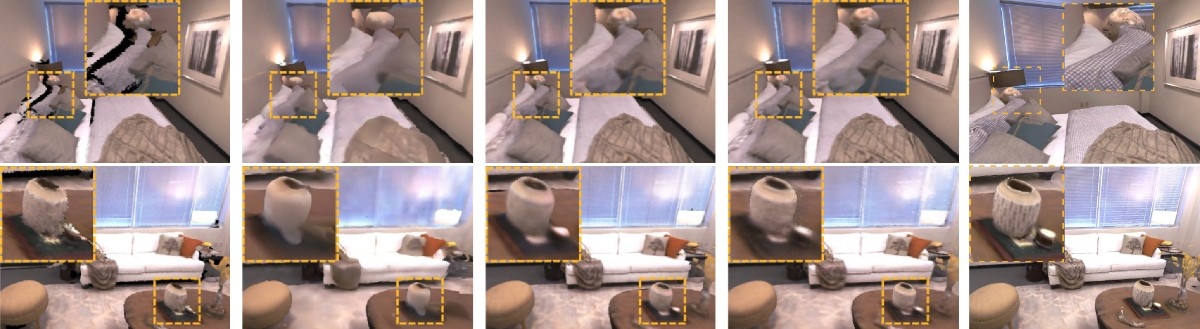}
            \put(10, -2){\makebox(0,0){\small \textbf{(a) BundleFusion
            \cite{dai2017bundlefusion}}}}
            \put(30, -2){\makebox(0,0){\small \textbf{(b) Nice-SLAM \cite{zhu2022niceslam}}}}
            \put(50, -2){\makebox(0,0){\small \textbf{(c) ESLAM \cite{johari2023eslam}}}}
            \put(70, -2){\makebox(0,0){\small \textbf{(d) Co-SLAM \cite{wang2023coslam}}}}
            \put(90, -2){\makebox(0,0){\small \textbf{(e) Ours}}}
        \end{overpic}
        \vspace{0.4em}
        \captionof{figure}{\textbf{Qualitative comparison of diverse SLAM systems on the Replica dataset.} As highlighted in the zoomed-in regions, existing methods often suffer from blurry artifacts or incomplete geometry. In contrast, our approach preserves finer details and high-frequency textures (e.g., the patterns on the vase and bedsheets), resulting in superior photorealism.}
        \label{fig:placeholder}
    \end{center}
}]%

\begingroup
\renewcommand{\thefootnote}{}%
\footnotetext{$^{1}$Cognitive Robotics Lab, Department of Electrical Engineering and Computer Science, University of Arkansas, USA.}%
\footnotetext{$^{2}$Center for AI Research, VinUniversity, Ha Noi, Vietnam.}%
\footnotetext{$^{*}$Corresponding author: \texttt{tuand@uark.edu}}%
\endgroup

\begin{abstract}
Recent 3D Gaussian Splatting (3DGS) has enabled efficient photorealistic view synthesis and is rapidly being adopted in simultaneous localization and mapping (SLAM) systems for online mapping. In these systems, a Gaussian map must be expanded and refined incrementally while tracking runs in real time, so initialization and density control directly determine where limited computation and iterations are spent. This contrasts with offline 3DGS reconstruction, where such heuristics can be amortized over long optimization schedules. However, most 3DGS-SLAM pipelines inherit initialization and density-control heuristics from offline reconstruction, which can become brittle under the strict per-keyframe optimization budgets and incremental map growth of online SLAM. In this work, we revisit these heuristics in a decoupled 3DGS-SLAM setting and propose three geometry-aware methods that operate in the mapping thread: transmittance-preserving densification, camera-aware scale initialization from depth and intrinsics, and error-guided densification that focuses new primitives on high-residual regions. Our results show consistent improvements in rendering quality with negligible overhead, highlighting the coupling between photometric residuals and pose uncertainty in online SLAM. We will open-source our code to the community to foster growth and validate reproducibility. 


\end{abstract}



\input{01_introduction}

\input{02_related_work}
\input{03_system}

\input{04_methodology}

\input{05_evaluation}

\input{06_conclusions}

\bibliographystyle{IEEEtran}
\bibliography{IEEEabrv, 09_references}

\end{document}

%% file: 01_introduction.tex
\section{Introduction}
Simultaneous Localization and Mapping (SLAM) is a core capability for robots and augmented reality (AR) devices: it provides geometric and camera-trajectory information, enabling navigation, interaction, and scene understanding \cite{cadena2016past, taketomi2017visual, mur2015orb, campos2021orbslam3}. Recently, the community has also pursued \emph{photorealistic} online mapping, where the map supports high-fidelity view synthesis in addition to localization.
A major catalyst is 3DGS ~\cite{kerbl2023gaussian}, which represents a scene as anisotropic Gaussians and renders them efficiently via differentiable rasterization, achieving high-quality real-time novel-view synthesis.
This has quickly led to 3DGS-based SLAM systems such as Photo-SLAM~\cite{huang2024photoslam}, MonoGS~\cite{matsuki2024gaussiansplattingslam}, SplaTAM~\cite{keetha2024splatam}, and GS-SLAM~\cite{yan2024gsslam}.

\del{Despite this momentum, most 3DGS-SLAM systems inherit key \emph{initialization} and \emph{density-control} heuristics from the original offline 3DGS pipeline~\cite{kerbl2023gaussian}, which are not designed for the strict optimization budgets and incremental map growth of online SLAM.}

Despite this momentum, most 3DGS-SLAM systems inherit key \emph{initialization} and \emph{density-control} heuristics from the original offline 3DGS pipeline~\cite{kerbl2023gaussian}. However, recent studies highlight that these original heuristics lead to uncontrolled primitive growth and systematic opacity biases even in offline scenarios~\cite{bulo2024revising}. Consequently, they are fundamentally ill-suited for the strict optimization budgets, real-time computational constraints, and incremental map growth inherent to online SLAM~\cite{matsuki2024gaussiansplattingslam, keetha2024splatam}. First, adaptive density control relies on \emph{cloning} Gaussians in under-reconstructed regions based on view-space gradients~\cite{kerbl2023gaussian}. However, repeated cloning systematically biases alpha compositing and causes opacity drift~\cite{bulo2024revising}. Specifically, because the original pipeline simply duplicates a primitive while preserving its original opacity, the resulting overlapping Gaussians artificially over-occlude the background during rendering~\cite{bulo2024revising}. This double-counting introduces a systematic bias that inflates the photometric contribution of newly cloned Gaussians, leading to localized opacity drift and hindering accurate optimization of the underlying scene structures. Second, purely gradient-triggered densification is \emph{passive}: with only hundreds of mapping iterations per keyframe in online SLAM (versus tens of thousands of iterations typical for offline reconstruction), persistently under-modeled regions may never accumulate gradients strong enough to attract new primitives. These effects are subtle but compounding, and they disproportionately affect online mapping quality.


In this paper, we revisit these heuristics under the online SLAM regime and propose geometry-aware methods that operate \emph{only} in the mapping thread. As shown in Fig.~\ref{fig:placeholder}, our method significantly increases the quality of existing SLAM benchmarks by accurately reconstructing high-frequency textures and fine geometric details. We build on the decoupled design of Photo-SLAM~\cite{huang2024photoslam}, where ORB-SLAM3~\cite{campos2021orbslam3} performs real-time tracking while a parallel thread optimizes a Gaussian map for photorealistic rendering. We observe that under strict iteration budgets, rendering quality is dominated by (1) \emph{where} new primitives are inserted, (2) \emph{how} their initial scale reflects camera geometry rather than point density, and (3) whether densification preserves the intended alpha-compositing behavior. We identify three offline 3DGS heuristics that are brittle in online SLAM and propose three simple, geometry-aware fixes that modify mapping only and introduce no learned components. Our contributions are :
\begin{itemize}
    \item We introduce \textbf{Transmittance-Preserving Densification} to preserve composite opacity under cloning, preventing systematic opacity inflation under alpha compositing during online densification.
    \item We introduce  \textbf{Camera-aware Scale initialization} to initialize Gaussian size from the metric pixel footprint at the observed depth (using camera intrinsics and depth).
    \item We introduce  \textbf{Error-guided Densification} to allocate new primitives to persistently high-residual image regions by sampling an error map and spawning Gaussians via depth-backed back-projection.
\end{itemize}

%% file: 02_related_work.tex
\section{Related Work}

\noindent\textbf{Neural rendering for dense SLAM.}
NeRF-based dense SLAM methods jointly optimize camera poses and a neural scene representation online~\cite{mildenhall2020nerf,sucar2021imap,zhu2022niceslam,wang2023coslam,johari2023eslam}.
While these implicit fields enable high-quality view synthesis, ray-based volume rendering and per-frame backpropagation remain expensive, so practical systems often trade off resolution by sampling only a subset of pixels/rays during optimization~\cite{sucar2021imap,zhu2022niceslam,wang2023coslam,johari2023eslam}. The 3DGS instead represents the scene with explicit anisotropic Gaussians and optimizes them through differentiable splatting, enabling efficient high-resolution rendering and gradients~\cite{kerbl2023gaussian}.
Recent 3DGS-SLAM systems differ mainly in how tightly tracking is coupled to Gaussian optimization:
MonoGS~\cite{matsuki2024gaussiansplattingslam}, SplaTAM~\cite{keetha2024splatam}, and GS-SLAM~\cite{yan2024gsslam} directly track against the Gaussian map (with different robustness and speed--accuracy strategies),
whereas Photo-SLAM~\cite{huang2024photoslam} follows a decoupled design with feature-based tracking running in parallel with an asynchronous photorealistic Gaussian mapping thread.
We adopt this decoupled backbone to isolate mapping-side improvements under real-time tracking constraints.

\textbf{Online Gaussian map growth: initialization, densification, and evaluation.}
The original 3DGS reconstruction pipeline initializes primitives from sparse points (with k-NN-style scale heuristics) and expands representation capacity through Adaptive Density Control (ADC), which clones/splits Gaussians based on view-space gradient statistics while pruning low-opacity primitives~\cite{kerbl2023gaussian}.
However, recent analyses show that naive cloning can bias alpha compositing by increasing opacity in repeatedly densified regions, and that purely gradient-threshold densification may fail to allocate capacity to persistently under-modeled high-frequency regions; error-driven density control and opacity-aware corrections can mitigate these issues~\cite{bulo2024revising}.
ConeGS~\cite{baranowski2025conegs} further explores error-guided primitive insertion and initializes sizes from the pixel viewing footprint (inspired by mip-NeRF-style cones/frustums~\cite{barron2021mipnerf}), \cite{kheradmand20243d}, but is studied primarily in offline reconstruction settings with auxiliary depth proxies. The ConeGS model is inspired by the iNGP architecture~\cite{muller2022instant}, adapting its efficiency to the 3D Gaussian Splatting (3D GS) framework.
In online SLAM, map growth is incremental and optimization per keyframe is tightly budgeted, making initialization and densification decisions disproportionately important.
Our work targets online RGB-D 3DGS-SLAM in this regime: keeping the decoupled SLAM backbone unchanged, we introduce geometry-aware, mapping-only methods that stabilize densification under alpha compositing, initialize scales from metric pixel footprints using depth and intrinsics, and allocate new primitives to persistently high-residual regions under short-horizon optimization.
Finally, because rendering metrics on training keyframes can be overly optimistic due to overfitting, we follow prior recommendations and report rendering quality along the full trajectory rather than keyframe-only train-view evaluation~\cite{keetha2024splatam}.

%% file: 03_system.tex




%% file: 04_methodology.tex
\begin{figure*}[ht]
    \centering
    \includegraphics[width=1\linewidth]{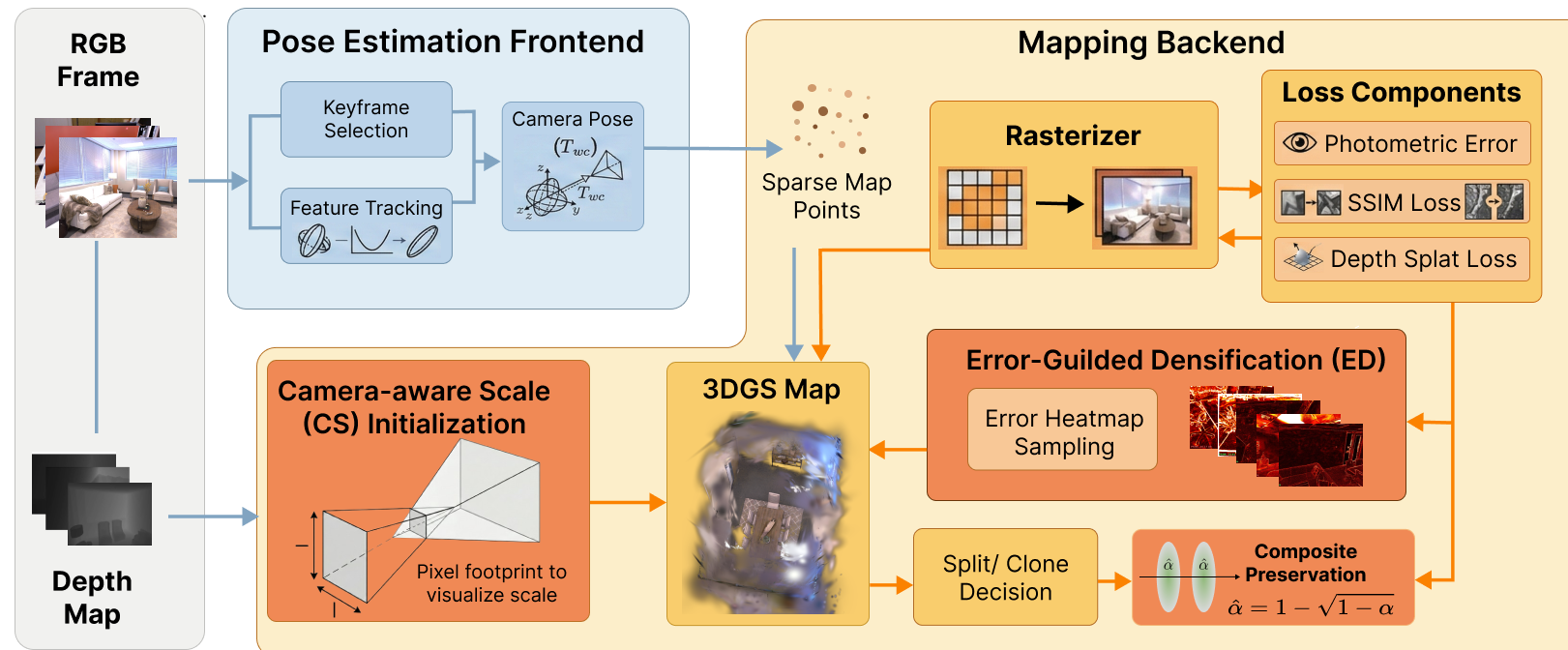}
    \caption{\textbf{System overview.} Our pipeline follows Photo-SLAM’s decoupled design: ORB-SLAM3 estimates poses and selects keyframes, while an asynchronous backend optimizes a 3D Gaussian Splatting map via differentiable rasterization. mapping modules: Camera-aware Scale initialization, Transmittance-Preserving Densification, and Error-guided Densification; the tracking frontend remains unchanged.
    }
    \label{fig:system}
\end{figure*}

\section{Methodology}
\label{sec:method}

\subsection{System Overview}

Fig.~\ref{fig:system} illustrates our decoupled 3DGS-SLAM pipeline. Building upon the Photo-SLAM framework~\cite{huang2024photoslam}, our system operates via two parallel threads. The real-time tracking frontend (ORB-SLAM3, RGB-D mode) processes incoming frames to estimate camera poses $\mathbf{T}_{wc}^t$ and select keyframes. Concurrently, the asynchronous mapping backend optimizes a 3DGS map by rendering from keyframe poses and minimizing a photometric reconstruction loss. While the tracking stack, rendering model, and mapping objective remain unchanged from the baseline, we redesign the density-control and initialization mechanisms to operate effectively under a tight online computational budget. Our mapping-only pipeline integrates three key modules directly into the optimization loop. The following sections detail the mathematical formulation of the baseline and our proposed modifications.

\subsection{Problem Formulation}
\label{sec:method_baseline}



We build on the decoupled architecture of Photo-SLAM~\cite{huang2024photoslam},
where a real-time tracking thread (ORB-SLAM3~\cite{campos2021orbslam3})
estimates camera poses, and a parallel mapping thread maintains a 3D Gaussian map
optimized through differentiable splatting.
Let $t \in \mathcal{K}$ index keyframes selected by the tracking frontend.
We denote the RGB observation by $I_t$,
camera intrinsics by $\mathbf{K}$,
and the world-from-camera pose by $\mathbf{T}^{t}_{wc} \in SE(3)$
(with $\mathbf{T}^{t}_{cw} = (\mathbf{T}^{t}_{wc})^{-1}$).

Following standard 3DGS~\cite{kerbl2023gaussian}, the map is a set of Gaussian primitives
\begin{equation}
\mathcal{G} = \{ g_i \}_{i=1}^{M},
\end{equation}
where each primitive $i$ is parameterized by a 3D mean $\boldsymbol{\mu}_i$,
an anisotropic covariance $\boldsymbol{\Sigma}_i$,
an opacity $\alpha_i$,
and appearance coefficients (e.g., spherical harmonics).
Given $\mathcal{G}$, $\mathbf{K}$, and a keyframe pose $\mathbf{T}^{t}_{cw}$, the differentiable splatting renderer $\mathcal{R}$ produces a rendered image: 
\begin{equation}
\hat{I}_t
=
\mathcal{R}\!\left(\mathcal{G}; \mathbf{T}^{t}_{cw}, \mathbf{K}\right).
\end{equation}
Mapping then minimizes a photometric reconstruction objective over keyframes
\begin{equation}
\mathcal{L}_{\mathrm{photo}}(\mathcal{G})
=
\sum_{t \in \mathcal{K}}
\left\|
\hat{I}_t - I_t
\right\|_1,
\end{equation}
interleaved with \emph{adaptive density control} that clones/splits Gaussians
based on view-space position gradients and prunes low-opacity primitives~\cite{kerbl2023gaussian}.
Abstractly, one mapping iteration can be written as


\begin{equation}
\mathcal{G}^{(k+\frac{1}{2})}
=
\mathcal{G}^{(k)}
-
\eta \nabla {\mathcal{G}}
\mathcal{L}_{\mathrm{photo}}(\mathcal{G}^{(k)}),
\end{equation}

\begin{equation}
\mathcal{G}^{(k+1)}
=
\mathcal{D}_{\star}\!\left(\mathcal{G}^{(k+\frac{1}{2})}\right),
\label{eq:density_control_operator}
\end{equation}
where $\mathcal{D}_{\star}$ denotes the density-control operator.
    
In Photo-SLAM~\cite{huang2024photoslam}, the decoupled design means that tracking provides $\{\mathbf{T}^{t}_{wc}\}_{t\in\mathcal{K}}$ in real time, and the mapping ${G}$ using Eqs.~(3)--(5)
with poses treated as fixed inputs. In offline 3DGS, initialization and density-control heuristics are amortized over long optimization schedules. In online SLAM, mapping is instead constrained by limited wall-clock compute.
In the Photo-SLAM implementation, each mapping iteration performs a single optimization step
on one selected active keyframe; hence, the effective number of updates received by each keyframe
is stochastic and run-dependent, rather than a fixed per-keyframe budget.
As a result, early design choices disproportionately affect where representational capacity is allocated
and how optimization gradients evolve over time.
In particular,
(i) neighbor-distance scale initialization ties Gaussian size to local point density
rather than scene geometry,
(ii) cloning operations can introduce systematic opacity inflation under alpha compositing,
and
(iii) purely gradient-triggered densification may fail to populate persistently under-modeled
regions within a tight iteration budget.

We observe that, under a tight online mapping budget, rendering quality is governed less by learning dynamics and more by three ``non-learning'' design choices:
\emph{where} new Gaussians are inserted,
\emph{how} they are initially scaled in metric space,
and whether density-control operations preserve the intended alpha-compositing behavior.
Accordingly, we introduce three contributions that operate in the mapping thread. Formally, we keep the tracking frontend (ORB-SLAM3), the renderer $\mathcal{R}(\cdot)$, and the objective $\mathcal{L}_{\mathrm{photo}}(\cdot)$ unchanged. We simply replace the baseline density-control operator $\mathcal{D}_{\mathrm{base}}$ in Eq.~\eqref{eq:density_control_operator} with our proposed operator $\mathcal{D}_{\mathrm{ours}}$. Both operators act on the space of Gaussian maps $\mathcal{G}$ where $\mathcal{D}_{\mathrm{ours}}$ differs from the baseline through the three mapping-side mechanisms introduced in Sec.~\ref{sec:method_ed}, Sec.~\ref{sec:method_cs}, and Sec.~\ref{sec:method_oc}.


\subsection{Error-guided Densification (ED)}
\label{sec:method_ed}

\begin{figure}[t]
    \centering
    \includegraphics[width=1\linewidth]{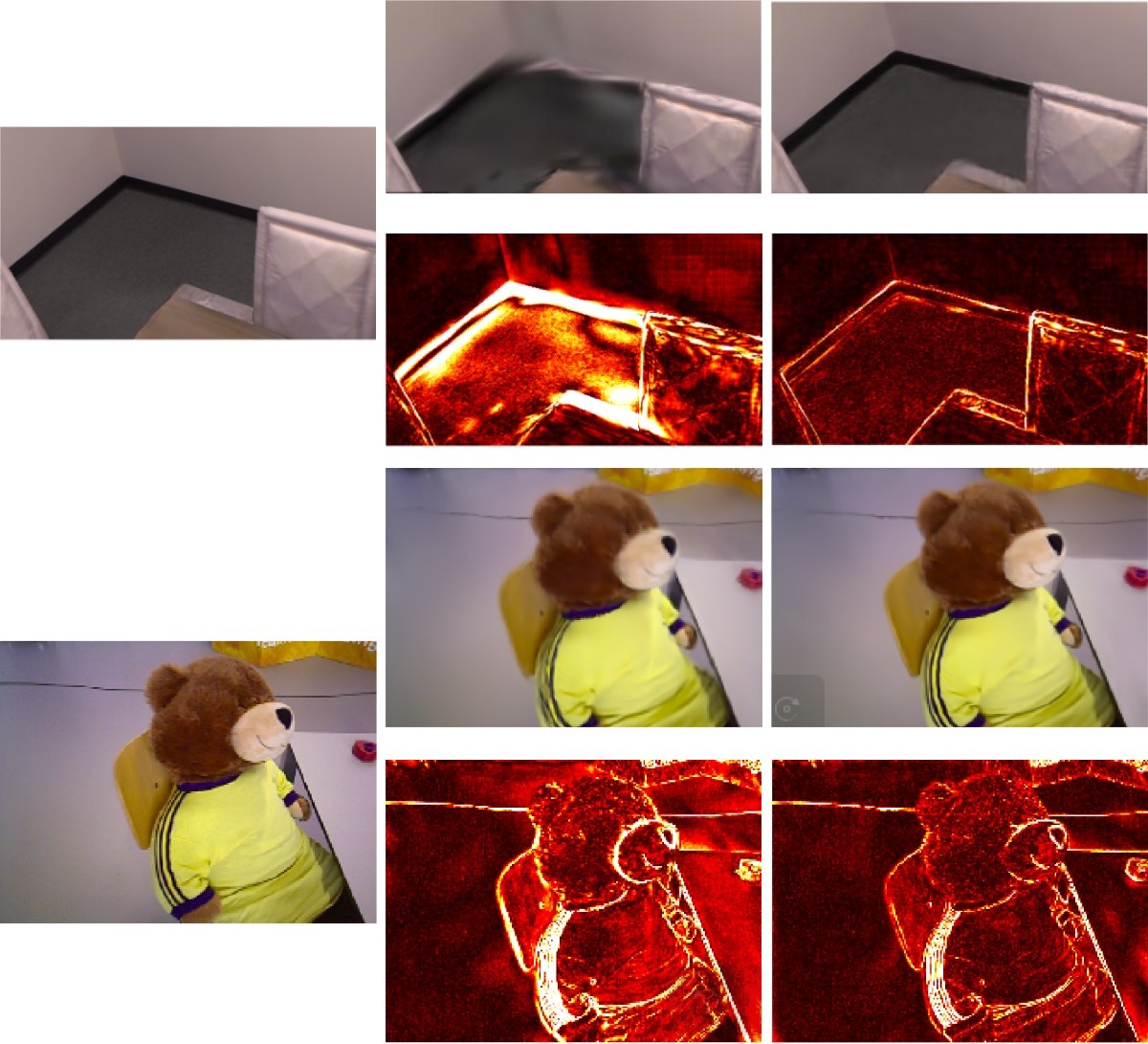}
    
    \vspace{2pt}
    \makebox[0.32\linewidth][c]{\textbf{a) Ground truth}} \hfill
    \makebox[0.32\linewidth][c]{\textbf{b) Photo-SLAM}} \hfill
    \makebox[0.32\linewidth][c]{\textbf{c) Ours}}
    \vspace{4pt}
    
    \caption{\textbf{Effect of Error-guided Densification under the same online mapping budget.} For each example, we show the observed keyframe (left), the baseline gradient-based densification (middle), and ours with ED (right). The residual heatmaps is brighter indicates larger error), where ED reduces holes and over-smoothed regions.
    }
    \label{fig:result_1}

\end{figure}

Standard 3DGS performs adaptive density control by cloning/splitting Gaussians whose \emph{average view-space positional gradient magnitude} exceeds a threshold~\cite{kerbl2023gaussian}. This signal is effective in offline reconstruction, where long optimization schedules allow photometric errors to propagate into gradients and eventually trigger densification near under-modeled regions. In online decoupled SLAM, however, each keyframe receives only a limited number of mapping updates (Sec.~\ref{sec:setup}), so regions with insufficient primitive coverage may not \emph{exhibit sufficiently large} view-space positional gradients early enough, leading to persistent holes or over-smoothed surfaces as shown in Fig.~\ref{fig:result_1}.



\textbf{Error-guided insertion.}
We therefore add an auxiliary densification operator that directly spawns new Gaussians at high-residual pixels.
Let keyframe $t$ provide an RGB observation $\mathbf{I}_t$ and a depth map $\mathbf{D}_t$ (RGB-D setting), with intrinsics $\mathbf{K}$
and a pose estimate $\mathbf{T}^t_{cw}\in SE(3)$ (world $\rightarrow$ camera) from the tracking frontend; we denote its inverse by $\mathbf{T}^t_{wc}=(\mathbf{T}^t_{cw})^{-1}$.
Rendering the current Gaussian map $\mathcal{G}$ at pose $\mathbf{T}^t_{cw}$ yields
$\hat{\mathbf{I}}_t=\mathcal{R}(\mathcal{G};\mathbf{T}^t_{cw},\mathbf{K})$.
Every $K_{\mathrm{ED}}$ mapping iterations, we compute the per-pixel photometric residual
\begin{equation}
e_t(u,v)=\left\|\hat{\mathbf{I}}_t(u,v)-\mathbf{I}_t(u,v)\right\|_{1}.
\label{eq:residual_map}
\end{equation}

\begin{equation}
\mathcal{V}_t=\left\{(u,v)\;\middle|\; e_t(u,v)>\tau,\; \mathbf{D}_t(u,v)\ \text{valid},\; g_t(u,v)<\delta\right\},
\label{eq:ed_candidates}
\end{equation}

where $g_t(u,v)$ is a depth-edge score used to suppress depth discontinuities, so that we avoid inserting Gaussians on depth edges. In practice we use a simple finite-difference filter,
$g_t(u,v)=\max\{|\mathbf{D}_t(u{+}1,v)-\mathbf{D}_t(u,v)|,\;|\mathbf{D}_t(u,v{+}1)-\mathbf{D}_t(u,v)|\}$.
We sample $N$ pixels $(u_i,v_i)\sim \mathrm{Cat}(\pi_t)$ from  candidate set $\mathcal{V}_t$ by importance sampling on the residual heatmap:
\begin{equation}
\pi_t(u,v)\propto e_t(u,v), \quad (u,v)\in\mathcal{V}_t .
\end{equation}

For each sampled pixel, we back-project using depth $d_i=\mathbf{D}_t(u_i,v_i)$:
\begin{equation}
\tilde{\mathbf{p}}_{c,i}=\bigl[\mathbf{p}_{c,i}^\top,\,1\bigr]^\top,\qquad
\tilde{\boldsymbol{\mu}}_i=\mathbf{T}^t_{wc}\tilde{\mathbf{p}}_{c,i},\qquad
\boldsymbol{\mu}_i=\tilde{\boldsymbol{\mu}}_{i,1:3}.
\label{eq:backproject}
\end{equation}

and spawn a new Gaussian with mean $\boldsymbol{\mu}_i$, color initialized from $\mathbf{I}_t(u_i,v_i)$, and a conservative opacity $\alpha_0$.
We initialize its covariance isotropically as $\boldsymbol{\Sigma}_i=s_{\text{cam}}(d_i)^2\mathbf{I}$ using the camera-aware scale rule in Sec.~\ref{sec:method_cs}, so that newly inserted primitives roughly match the pixel footprint at depth $d_i$.

 $\mathbf{p}_{c,i} \in \mathbb{R}^3$ is the 3D point in camera coordinates obtained by back-projecting sampled pixel $(u_i, v_i)$ using the sensor depth $d_i = \mathbf{D}_t(u_i, v_i)$: $$\mathbf{p}_{c,i} = d_i \cdot \mathbf{K}^{-1}\begin{bmatrix}u_i \\ v_i \\ 1\end{bmatrix}
 = \begin{bmatrix}d_i(u_i - c_x)/f_x \\ d_i(v_i - c_y)/f_y \\ d_i\end{bmatrix}$$

Step-by-step:
1. $\mathbf{p}_{c,i}$: 3D point in **camera frame** from depth back-projection
2. $\tilde{\mathbf{p}}_{c,i}$: homogeneous form of $\mathbf{p}_{c,i}$ (append 1)
3. $\tilde{\boldsymbol{\mu}}_i = \mathbf{T}^t_{wc}\,\tilde{\mathbf{p}}_{c,i}$: transform to **world frame** (4-vector)
4. $\boldsymbol{\mu}_i = \tilde{\boldsymbol{\mu}}_{i,\,1:3}$: extract the first 3 components as the new Gaussian's world-space center

ED complements (rather than replaces) the standard gradient-based clone/split operator by injecting primitives in image-space error regions that may not yet be supported by nearby Gaussians under tight online budgets.
All ED hyperparameters $(K_{\mathrm{ED}}, N, \tau, \delta, \alpha_0)$ are fixed across scenes and reported in Sec.~\ref{sec:setup}.


\subsection{Camera-aware Scale (CS) Initialization}
\label{sec:method_cs}

\begin{figure}
    \centering
    \includegraphics[width=1.0\linewidth]{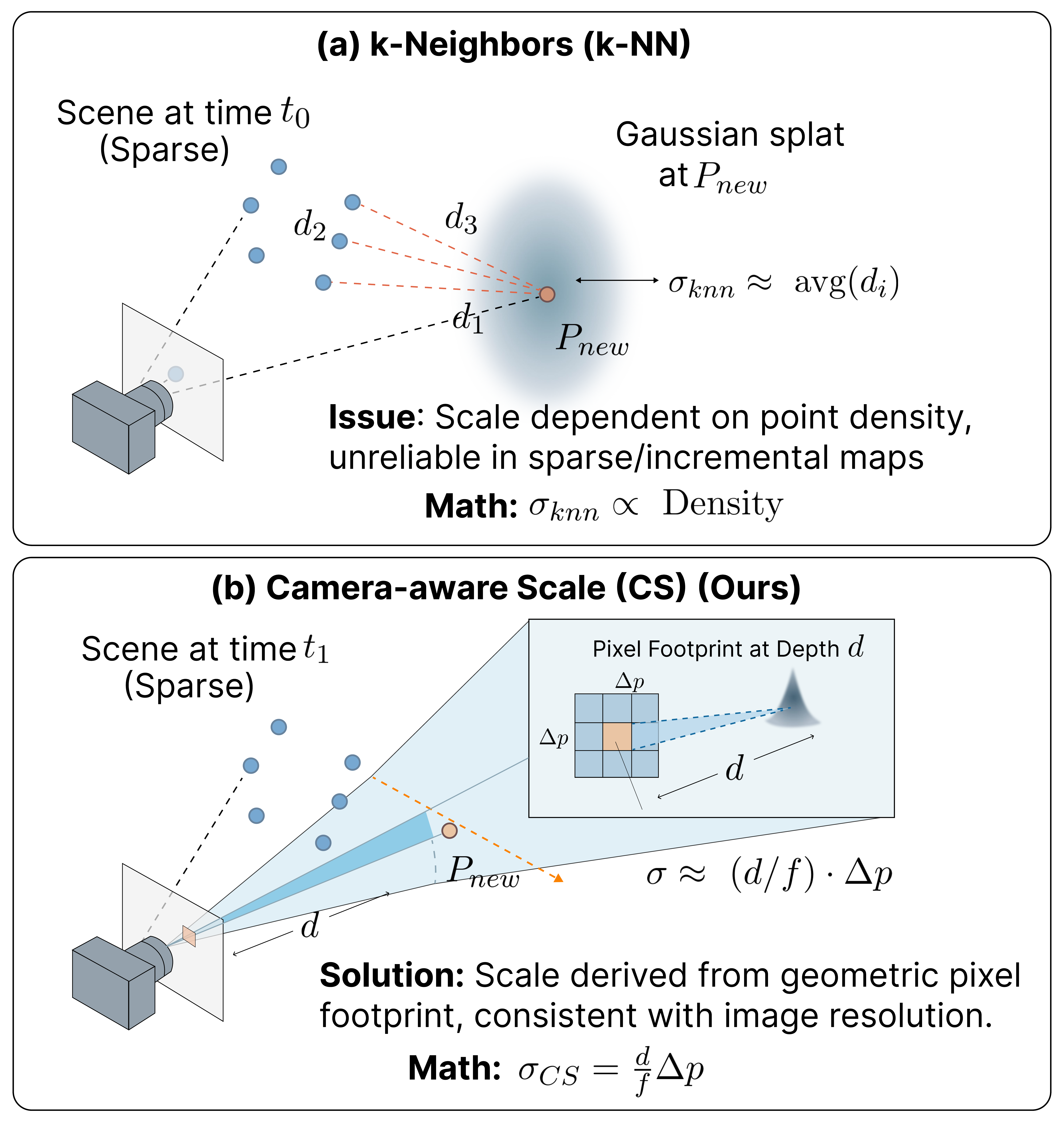}
    \caption{\textbf{Scale initialization in online SLAM.} (a) k-NN scale initialization depends on sparse point density, often producing considerable splats and blurry artifacts. (b) Camera-aware Scale initializes the Gaussian size from the metric pixel footprint at the observed depth, yielding consistent, sharper primitives independent of point density.
    }
    
    \label{fig:scale_comparison}
\end{figure}

3DGS initializes each Gaussian's covariance isotropically using a k-NN heuristic (e.g., the mean distance to its three nearest neighbors in an initial point cloud)~\cite{kerbl2023gaussian}.
This ties the initial scale to local point density, which is reasonable when the point cloud is dense and uniformly sampled.
In incremental SLAM, however, new points arrive sparsely and unevenly due to feature tracking and keyframe selection, so point density can reflect texture and pipeline stochasticity rather than scene geometry \cite{tran2025triags}. Under a short online optimization budget, inaccurate initial scales are harder to correct and can manifest as considerable splats (blurry surfaces) or undersized primitives (holes), as illustrated in Fig.~\ref{fig:scale_comparison}.

We replace density-based scale initialization with a camera-aware rule derived from the metric footprint of a pixel at depth $d$.
For a pinhole camera with focal lengths $(f_x,f_y)$ (in pixels), the back-projection of pixel $(u,v)$ with depth $d$ is
$\mathbf{p}_c=d\,\mathbf{K}^{-1}[u,v,1]^\top$.
A one-pixel change in image coordinates corresponds to metric displacements
$\Delta_x(d)=d/f_x$ and $\Delta_y(d)=d/f_y$ in camera space.
We define an isotropic initial scale proportional to this pixel footprint:
\begin{equation}
s_{\text{cam}}(d)=\lambda\,\frac{d}{\bar{f}}, \qquad \bar{f}=\tfrac{1}{2}(f_x+f_y),
\label{eq:cone}
\end{equation}
where $\lambda$ is a global coverage factor.
We initialize newly inserted Gaussians with $\boldsymbol{\Sigma}_0=s_{\text{cam}}(d)^2\mathbf{I}$ and then optimize anisotropic covariance as in standard 3DGS.
In our RGB-D setting, $d$ is directly available from the sensor depth  
at insertion time (e.g., for ED in Sec.~\ref{sec:method_ed}).

Eq.~\eqref{eq:cone} follows the same pixel-footprint intuition used by mip-NeRF-style conical frustums~\cite{barron2021mipnerf} and recent error-guided Gaussian insertion strategies~\cite{baranowski2025conegs},
but we apply it in an online RGB-D SLAM backend and use directly observed depth to set the initial scale at insertion time (rather than relying on an auxiliary depth predictor). All hyperparameters, including $\lambda$, are reported in Sec.~\ref{sec:setup}.

\subsection{Transmittance-Preserving Densification (TPD)}
\label{sec:method_oc}

Adaptive density control (ADC) in 3DGS increases representation capacity by \emph{cloning} Gaussians that exhibit large positional gradients~\cite{kerbl2023gaussian}. 
In the standard implementation, the cloned Gaussian inherits the parent's opacity. Recent analysis shows that this choice introduces a systematic bias under the alpha-compositing used by Gaussian splatting, effectively \emph{inflating opacity} in regions that undergo repeated cloning and reducing the contribution of farther primitives~\cite{bulo2024revising}.

\textbf{Opacity inflation under cloning.}
Consider the setting used in prior work~\cite{bulo2024revising}: rendering the \emph{center pixel} of a splat, where the primitive induces an alpha coefficient $\alpha_i \in (0,1)$. Let $T_{\mathrm{pre}}$ denote the accumulated transmittance from all primitives in front of $g_i$ along this pixel ray. Before cloning, the residual transmittance passed to all farther primitives is
$T_{\mathrm{pre}}(1-\alpha_i)$.
If we na\"ively clone $g_i$ into two overlapping copies with the same opacity, alpha compositing yields residual transmittance
$T_{\mathrm{pre}}(1-\alpha_i)^2$, which is strictly smaller for $\alpha_i\in(0,1)$ and thus over-weights the cloned region.

\textbf{Transmittance-preserving opacity correction.}
To prevent systematic opacity inflation \cite{rota2024revising}, when $g_i$ is cloned we overwrite the opacity of \emph{both} the parent and the clone with a corrected value $\hat{\alpha}_i$ such that the composite transmittance is preserved in the worst-case overlap scenario:
\[
T_{\mathrm{pre}}(1-\hat{\alpha}_i)^2 = T_{\mathrm{pre}}(1-\alpha_i).
\]
Solving gives the closed form
\begin{equation}
\hat{\alpha}_i = 1 - \sqrt{1-\alpha_i}.
\label{eq:opacity_correction}
\end{equation}
This correction is exact for the center-pixel overlap analysis, and, as argued in~\cite{bulo2024revising}, it reduces the cloning-induced bias for general pixels even though it cannot eliminate it completely for all screen-space overlap configurations.
We apply Eq.~\eqref{eq:opacity_correction} only to \emph{clone} operations; split offspring are displaced and resized, so the perfect-overlap assumption does not hold.


%% file: 05_evaluation.tex
\section{Evaluations}
\label{sec:eval}

\begin{figure}[h]
    \centering
    \begin{overpic}[width=\linewidth]{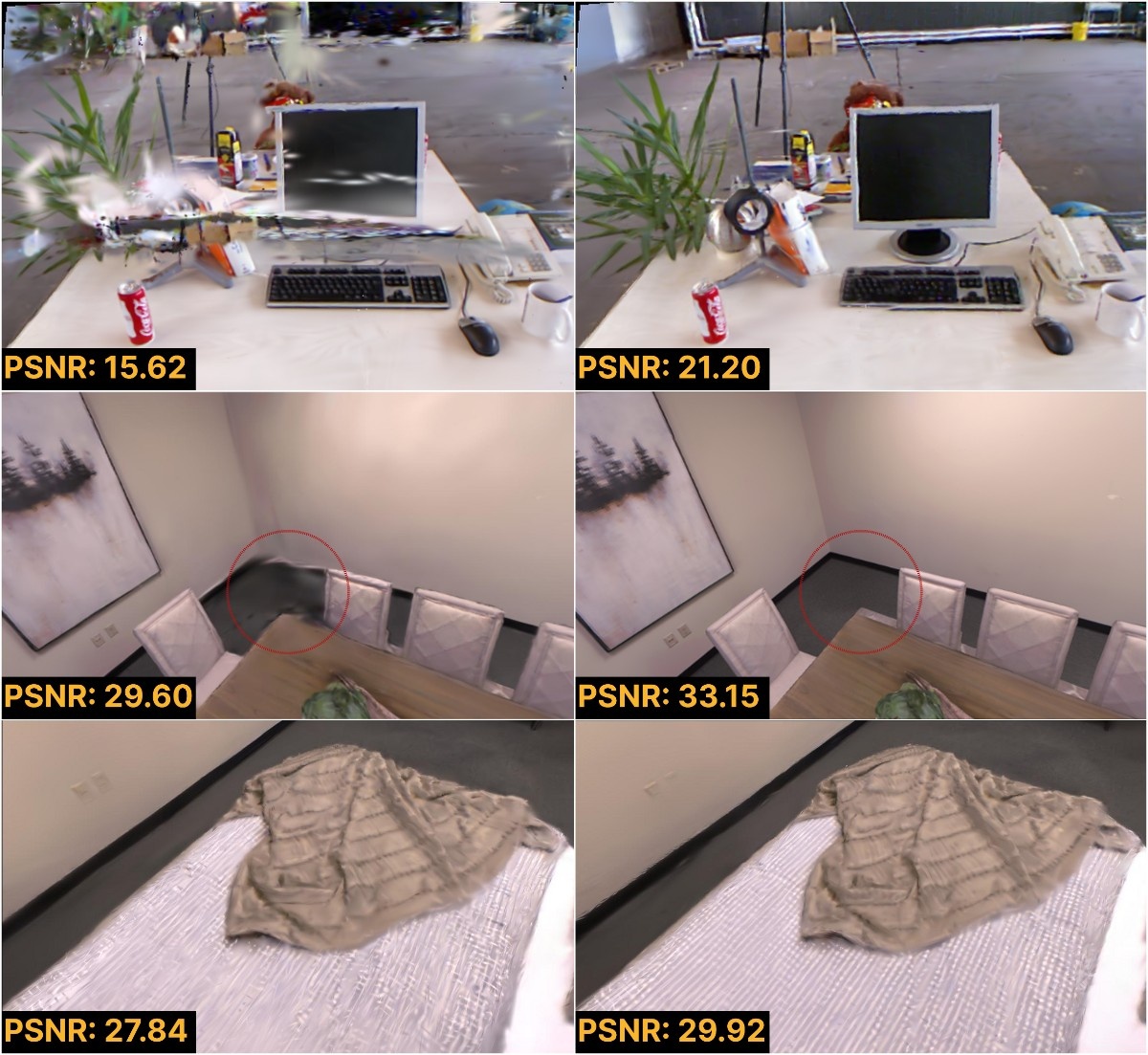}
        \put(24, -3){\makebox(0,0){\small \textbf{(a) Photo-SLAM \cite{huang2024photoslam}}}}
        \put(72, -3){\makebox(0,0){\small \textbf{(b) Our Rendering}}}
    \end{overpic}
    \vspace{0.5em}
    \caption{Comparison over representative views between (a) the Photo-SLAM baseline \cite{huang2024photoslam} and (b) our geometry-aware online mapping; PSNR (dB) is overlaid on the renderings.}
    \label{fig:comparison}
\end{figure}

\begin{figure}[h]
    \centering
    \begin{overpic}[width=\linewidth]{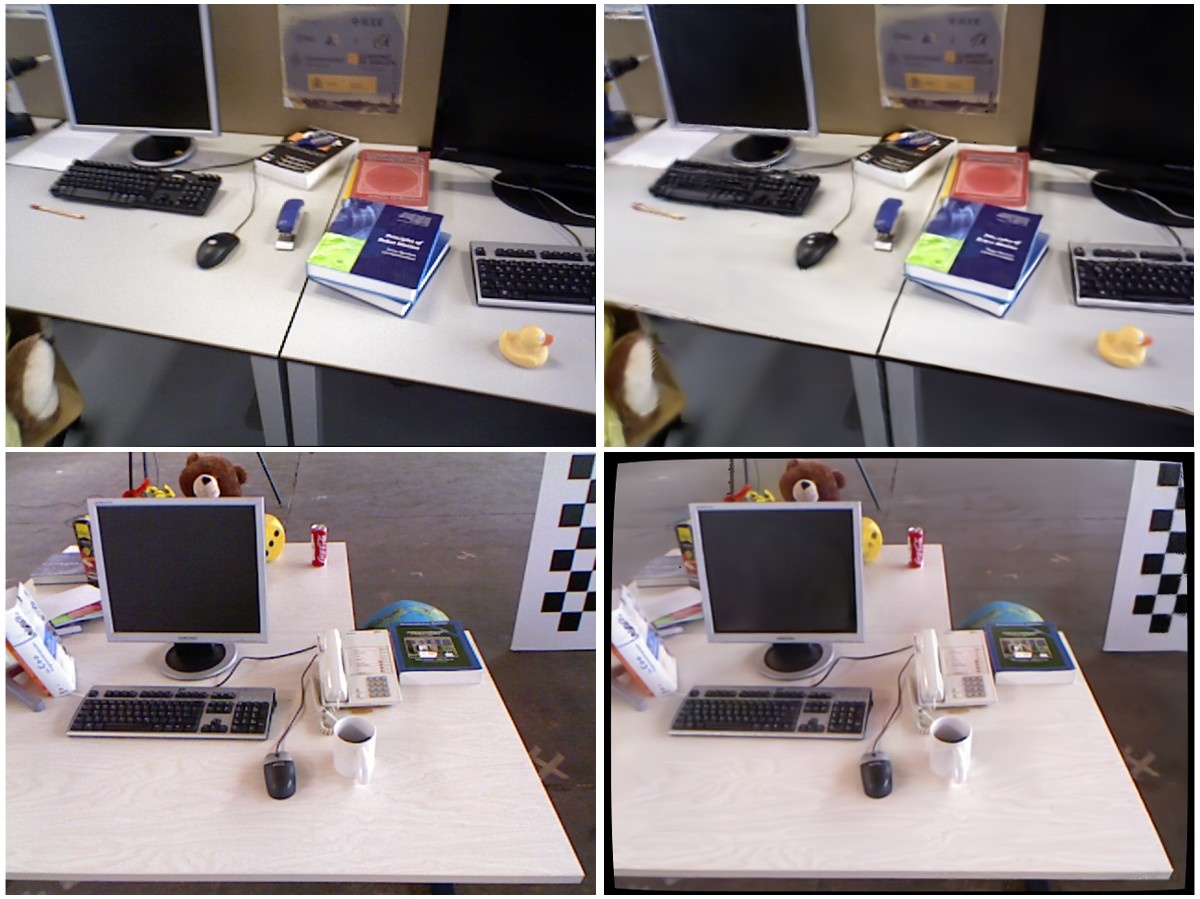}
        \put(25, -3){\makebox(0,0){\small \textbf{(a) Ground Truth}}}
        \put(75, -3){\makebox(0,0){\small \textbf{(b) Ours}}}
    \end{overpic}
    \vspace{0.5em}
    \caption{\textbf{Additional qualitative results on TUM RGB-D.} Top row: \textit{fr3/office}. Bottom row: \textit{fr1/desk}. Left: Ground Truth rendering. Right: Ours, demonstrating a high visual fidelity that closely matches the ground truth under the same online mapping budget.
    }
    \label{fig:tum_results}
\end{figure}

\begin{table*}[h]
\centering
\caption{\textbf{Rendering quality and tracking efficiency.} Evaluated on the full estimated trajectory (average of 3 runs). Our geometry-aware mapping consistently outperforms the baseline in rendering quality (PSNR, SSIM, LPIPS) across both datasets, while strictly maintaining real-time tracking performance (FPS). Best results are in \textbf{bold}.
}
\label{tab:render_efficiency}
\scriptsize
\setlength{\tabcolsep}{12.0pt}
\resizebox{\linewidth}{!}{
\begin{tabular}{c|l|cccc|cccc}
\toprule
& \textbf{Scene}
  & \multicolumn{4}{c|}{\textbf{Baseline (Photo-SLAM)}}
  & \multicolumn{4}{c}{\textbf{Ours}} \\
\cmidrule(lr){3-6}\cmidrule(lr){7-10}
& & PSNR$\uparrow$ & SSIM$\uparrow$ & LPIPS$\downarrow$ & FPS$\uparrow$
  & PSNR$\uparrow$ & SSIM$\uparrow$ & LPIPS$\downarrow$ & FPS$\uparrow$ \\
\midrule
\multirow{4}{*}{\rotatebox[origin=c]{90}{\textbf{TUM RGB-D}}}
& fr1/desk
  & 20.87 & .743 & .239 & 58.4
  & \best{21.44} & \best{.761} & \best{.216} & \best{79} \\
& fr2/xyz
  & 22.09 & .765 & .169 & 52.9
  & \best{23.27} & \best{.775} & \best{.145} & \best{72} \\
& fr3/office
  & \best{22.74} & \best{.780} & .154 & 43.7
  & 22.70 & .774 & \best{.152} & \best{52} \\
\cmidrule(lr){2-10}
& \textbf{Average}
  & 21.90 & .763 & .187 & 51.6
  & \best{22.47} & \best{.770} & \best{.171} & \best{68} \\
\midrule
\multirow{9}{*}{\rotatebox[origin=c]{90}{\textbf{Replica}}}
& office0
  & \best{38.48} & .964 & .050 & 48.6
  & 38.41 & \best{.968} & \best{.046} & \best{58} \\
& office1
  & 39.09 & .961 & .047 & 47.3
  & \best{39.61} & \best{.965} & \best{.041} & \best{62} \\
& office2
  & 33.03 & .938 & .077 & 44.1
  & \best{33.07} & \best{.944} & \best{.073} & \best{57} \\
& office3
  & \best{33.79} & .938 & .066 & 40.6
  & 33.63 & \best{.942} & \best{.063} & \best{54} \\
& office4
  & \best{36.02} & .952 & .054 & 39.9
  & 35.76 & \best{.955} & \best{.051} & \best{50} \\
& room0
  & 30.72 & .899 & .075 & 39.8
  & \best{31.76} & \best{.923} & \best{.066} & \best{53} \\
& room1
  & 33.51 & .934 & .057 & 43.4
  & \best{33.97} & \best{.943} & \best{.052} & \best{56} \\
& room2
  & 35.03 & .951 & .043 & 36.2
  & \best{35.49} & \best{.959} & \best{.039} & \best{49} \\
\cmidrule(lr){2-10}
& \textbf{Average}
  & 34.96 & .942 & .059 & 42.5
  & \best{35.21} & \best{.950} & \best{.054} & \best{55} \\
\bottomrule
\end{tabular}}
\end{table*}

\begin{table}[h]
\centering
\caption{Hyperparameters used by our method.}
\label{tab:ours_hparams}
\setlength{\tabcolsep}{2 pt} 
\resizebox{\linewidth}{!}{
\begin{tabular}{lcl}
\toprule
Symbol & Value & Description \\
\midrule
$\lambda$ & 1.0 & CS coverage factor in Eq.~\eqref{eq:cone} \\
$K_{\mathrm{ED}}$ & 100 & ED update period (mapping iterations) \\
$N$ & $0.5\,|\mathcal{P}_{\mathrm{err}}|$ & ED samples per update (50\% of error pixels) \\
$\tau$ & 0.02 & residual threshold for ED candidate pixels \\
$\delta$ & 0.5 & depth-edge threshold (finite differences) \\
$\alpha_0$ & 0.1 & initial opacity for ED-inserted Gaussians \\
\bottomrule
\end{tabular}}
\end{table}

\begin{table}[h]
\centering
\caption{Comparison with prior systems on Replica~\cite{straub2019replica}, averaged over 8 scenes. Results marked with $^\dagger$ are from prior work, while $^\ddagger$ indicates our results.}
\label{tab:sota}
\small
\setlength{\tabcolsep}{6pt}
\renewcommand{\arraystretch}{1.08}
\begin{tabular*}{\linewidth}{@{\extracolsep{\fill}}lccc@{\hspace{10pt}}c@{\hspace{4pt}}}
\toprule
\textbf{Method} & \textbf{PSNR}$\uparrow$ & \textbf{SSIM}$\uparrow$ & \textbf{LPIPS}$\downarrow$ & \textbf{FPS}$\uparrow$ \\
\midrule
BundleFusion$^\dagger$~\cite{dai2017bundlefusion} & 23.84 & 0.822 & 0.197 & 8.6 \\
Nice-SLAM$^\dagger$~\cite{zhu2022niceslam} & 26.16 & 0.832 & 0.232 & 2.3 \\
ESLAM$^\dagger$~\cite{johari2023eslam} & 30.59 & 0.866 & 0.162 & 6.7 \\
Co-SLAM$^\dagger$~\cite{wang2023coslam} & 30.25 & 0.864 & 0.175 & 14.6 \\
\midrule
SplaTAM$^\dagger$~\cite{keetha2024splatam} & 34.11 & 0.970 & 0.100 & $\sim$0.5 \\
GS-SLAM$^\dagger$~\cite{yan2024gsslam} & 34.27 & \textbf{0.975} & 0.082 & $\sim$8.0 \\
\midrule
Photo-SLAM$^\dagger$~\cite{huang2024photoslam} & 34.96 & 0.942 & 0.059 & 42.5 \\
\textbf{Ours}$^\ddagger$ & \textbf{35.21} & 0.950 & \textbf{0.054} & \textbf{55.0} \\
\bottomrule
\end{tabular*}
\end{table}

\begin{table}[h]
\centering
\caption{Tracking accuracy (ATE RMSE in cm, mean~$\pm$~std over 3 runs).
  TPD and CS are mapping-only; differences from Baseline lie within
  ORB-SLAM3 run-to-run variance.
  }
\label{tab:tracking}
\setlength{\tabcolsep}{12 pt} 
\resizebox{\linewidth}{!}{
\begin{tabular}{c|l cc}
\toprule
\textbf{Dataset} & \textbf{Scene} & \textbf{Baseline} & \textbf{Ours} \\
\midrule
\multirow{4}{*}{\rotatebox[origin=c]{90}{\textbf{TUM}}}
& fr1/desk     & $1.53\pm0.01$ & $1.53\pm0.01$ \\
& fr2/xyz      & $0.31\pm0.02$ & $0.30\pm0.01$ \\
& fr3/office   & $0.90\pm0.02$ & $0.94\pm0.02$ \\
\cmidrule{2-4}
& \textit{Average} & $\mathbf{0.91}$ & $0.92$ \\
\midrule
\multirow{9}{*}{\rotatebox[origin=c]{90}{\textbf{Replica}}}
& office0      & $0.45\pm0.02$ & $0.45\pm0.02$ \\
& office1      & $0.44\pm0.09$ & $0.37\pm0.02$ \\
& office2      & $1.04\pm0.17$ & $0.98\pm0.10$ \\
& office3      & $0.38\pm0.01$ & $0.38\pm0.02$ \\
& office4      & $0.50\pm0.03$ & $0.53\pm0.08$ \\
& room0        & $0.31\pm0.03$ & $0.32\pm0.01$ \\
& room1        & $0.39\pm0.01$ & $0.35\pm0.01$ \\
& room2        & $0.20\pm0.00$ & $0.21\pm0.01$ \\
\cmidrule{2-4}
& \textit{Average} & $0.46$ & $\mathbf{0.45}$ \\
\midrule
\multicolumn{2}{c|}{\textit{Overall Average}} & $0.59$ & $\mathbf{0.58}$ \\
\bottomrule
\end{tabular}}
\end{table}

\begin{table}[h]
\centering
\caption{Rendering-quality ablation within our method (averages only), mean of 3 runs. Rendering metrics via Photo-SLAM-eval.}
\label{tab:render_ablation_ours_quality_avg}
\scriptsize
\setlength{\tabcolsep}{3 pt}
\resizebox{\linewidth}{!}{ 
\begin{tabular}{l|ccc|ccc}
\toprule
\multirow{2}{*}{\textbf{Dataset}} & \multicolumn{3}{c|}{\textbf{+TPD+CS}} &
\multicolumn{3}{c}{\textbf{+TPD+CS+ED}} \\
\cmidrule(lr){2-4}\cmidrule(lr){5-7}
& PSNR$\uparrow$ & SSIM$\uparrow$ & LPIPS$\downarrow$
& PSNR$\uparrow$ & SSIM$\uparrow$ & LPIPS$\downarrow$ \\
\midrule
TUM RGB-D & 21.91 & .771 & .182 & 21.85 & .762 & .184 \\
Replica   & 34.72 & .944 & .055 & 34.39 & .931 & .055 \\
\midrule
\textbf{Overall Avg} & 31.22 & .891 & .082 & 30.97 & .874 & .083 \\
\bottomrule
\end{tabular}}
\end{table}

\begin{table}[h]
\centering
\caption{Iteration ablation on TUM RGB-D (average of fr1/desk and fr2/xyz). Performance of the baseline versus our proposed mapping modifications under varying per-keyframe iteration budgets.}
\label{tab:iter_ablation}

\setlength{\tabcolsep}{3 pt}
\resizebox{\linewidth}{!}{ 
\begin{tabular}{c|cc|cc|cc|cc}
\toprule
\multirow{2}{*}{\textbf{Iters / KF}} & \multicolumn{2}{c|}{\textbf{PSNR $\uparrow$}} & \multicolumn{2}{c|}{\textbf{SSIM $\uparrow$}} & \multicolumn{2}{c|}{\textbf{LPIPS $\downarrow$}} & \multicolumn{2}{c}{\textbf{\#Gaussians (K)}} \\
& Base & Ours & Base & Ours & Base & Ours & Base & Ours \\
\midrule
50  & 13.70 & \textbf{13.80} & 0.575 & \textbf{0.578} & 0.598 & \textbf{0.588} & 3.50 & 3.50 \\
100 & 12.90 & \textbf{14.25} & 0.575 & \textbf{0.582} & 0.588 & \textbf{0.580} & 1.90 & 1.90 \\
200 & \textbf{15.90} & 15.70 & 0.613 & \textbf{0.618} & 0.498 & \textbf{0.490} & 2.30 & 2.40 \\
300 & \textbf{16.90} & 16.80 & \textbf{0.638} & 0.635 & \textbf{0.425} & 0.428 & 2.85 & 2.83 \\
\bottomrule
\end{tabular}%
}
\end{table}

\subsection{Experimental Setup}
\label{sec:eval_setup}

We evaluate our mapping-side modifications along two axes: (i) \emph{photorealistic rendering quality} of the final Gaussian map when rendered from the estimated camera trajectory, and (ii) \emph{camera tracking accuracy}. We compare three configurations: (1) the Photo-SLAM baseline~\cite{huang2024photoslam}, (2) baseline +Transmittance-Preserving Densification + Camera-aware Scale initialization (+TPD+CS), and (3) baseline +TPD+CS + Error-guided Densification (+ED). Importantly, TPD/CS/ED modify the \emph{mapping} thread only; tracking (ORB-SLAM3) is unchanged.

We follow Photo-SLAM's RGB-D setting and evaluate on two standard RGB-D benchmarks: TUM RGB-D ~\cite{sturm2012tum} \del{ (3 sequences: \texttt{fr1/desk}, \texttt{fr2/xyz}, \texttt{fr3/long\_office\_household})}
and Replica \cite{straub2019replica}. \del{ (8 synthetic scenes: \texttt{office0--4}, \texttt{room0--2}),
for a total of 11 scenes.}
All experiments are conducted on an NVIDIA RTX 5090 (32GB) with CUDA 12.8. We use the official Photo-SLAM evaluation toolkit~\cite{photoslam_eval} to compute PSNR, SSIM~\cite{wang2004ssim}, and LPIPS~\cite{zhang2018lpips}.
Crucially, the toolkit renders the saved Gaussian model along the \emph{entire} estimated trajectory by timestamp-associating poses with the dataset RGB stream, rather than evaluating only on training keyframes~\cite{photoslam_eval}. This choice is deliberate: prior work has noted that reporting rendering metrics on the same input (training) views can be misleading because high-capacity models may simply overfit; evaluating on held-out/novel views is more informative~\cite{keetha2024splatam}. While standard SLAM benchmarks do not always provide fully independent hold-out views, rendering on the full trajectory (including non-keyframes) provides a substantially more demanding and honest proxy than keyframe-only evaluation.

SLAM systems are inherently non-deterministic due to multi-thread scheduling and feature-level stochasticity. We therefore run each configuration three times per scene and report mean$\pm$std to ensure fairness. This practice is also consistent with Photo-SLAM's own recommendation to repeat runs to reduce the impact of nondeterminism~\cite{huang2024photoslam}. Unless otherwise specified, we follow the official Photo-SLAM implementation and keep all baseline tracking and mapping parameters unchanged. For mapping, we use the provided Gaussian-mapper YAML configurations for each dataset, which fix the optimizer and the standard 3DGS density-control schedule (e.g., a densification interval of 100 iterations and a gradient threshold of $0.001$; dataset-specific pruning/densification ranges are kept as in the configs). Our mapping-side modifications introduce only a small set of additional hyperparameters summarized in Table~\ref{tab:ours_hparams}, and we keep them fixed across all scenes.
\label{sec:setup}

OC is parameter-free: when cloning is triggered by the baseline density controller, we apply the closed-form opacity correction in Eq.~\eqref{eq:opacity_correction}.
Camera-aware Scale initialization uses a single coverage factor $\lambda$ in Eq.~\eqref{eq:cone}.
Error-guided Densification (ED) is executed every $K_{\mathrm{ED}}$ mapping iterations; at each ED step as shown in Table \ref{tab:render_ablation_ours_quality_avg},
we sample $N$ pixels from the residual map using threshold $\tau$, filter depth edges using $\delta$,
and initialize spawned Gaussians with opacity $\alpha_0$ (Sec.~\ref{sec:method_ed}).

\subsection{Rendering Quality}
\label{sec:eval_rendering}
Table~\ref{tab:render_efficiency} reports full-trajectory rendering quality over all 11 scenes. Across both the real-world TUM RGB-D and synthetic Replica benchmarks, our geometry-aware mapping (TPD + CS) achieves the best average PSNR, improving from 21.90 to 22.47~dB on TUM and from 34.96 to 35.21~dB on Replica, our geometry-aware online mapping significantly improves rendering quality over the Photo-SLAM \cite{huang2024photoslam} baseline, achieving higher PSNR scores across representative views and effectively eliminating artifacts in complex regions (red circles) as shown in Fig.~\ref {fig:comparison}. Although PSNR gains are not uniform across all sequences, the overall trend is supported by corresponding improvements in average SSIM and reductions in LPIPS. The largest PSNR gains are observed on more challenging scenes such as fr2/xyz (+1.18~dB) and room0 (+1.04~dB), indicating that the proposed design is particularly beneficial under difficult online mapping conditions.

Furthermore, as demonstrated in Fig.~\ref{fig:tum_results}, our method is capable of recovering highly intricate details across diverse scenes such as fr3/office and fr1/desk. Notably, in well-observed regions, specific rendered frames can achieve a visual fidelity that closely approximates the ground truth, even under a strict online mapping budget.

\subsection{Efficiency Analysis}
\label{sec:eval_efficiency}


As shown in Table~\ref{tab:render_efficiency}, TPD/CS/ED introduces no measurable overhead in tracking throughput (FPS). Furthermore, Table~\ref{tab:tracking} confirms that our geometry-aware enhancements do not compromise tracking accuracy. Because our modifications operate strictly within the asynchronous mapping backend, the minor fluctuations in ATE RMSE between the baseline and our method are purely attributable to the inherent run-to-run stochasticity of the ORB-SLAM3 frontend.

Per-keyframe rendering time during mapping remains sub-millisecond across configurations. TPD+CS moderately increases the final number of Gaussians compared to the baseline. Intuitively, changing the initial scale distribution affects subsequent pruning/densification dynamics in adaptive density control. ED adds additional Gaussians in proportion to the number of spawning events and samples, as expected, but does not reduce the tracking FPS.

\textbf{Impact of Optimization Budget.} As shown in Table~\ref{tab:iter_ablation}, both the baseline and our method (+TPD+CS) converge to comparable rendering quality given a large budget (200--300 iterations per keyframe). However, under computational constraints of online SLAM (e.g., 100 iterations), the baseline suffers a severe performance drop. In contrast, our geometry-aware approach maintains stable map growth and superior photometric accuracy, demonstrating significantly faster convergence and robustness in resource-constrained settings.

\subsection{Context with Other 3DGS-SLAM Methods}
\label{sec:eval_context}

Table~\ref{tab:sota} provides context against published Gaussian-SLAM systems. Direct numerical comparison must be treated with caution because evaluation protocols differ substantially across papers---in particular, many works report rendering on training views, whereas our primary results follow a full-trajectory protocol via the Photo-SLAM evaluation toolkit~\cite{photoslam_eval}, motivated by concerns about overfitting and the limited informativeness of train-view rendering~\cite{keetha2024splatam}. Finally, our modifications preserve Photo-SLAM's core advantage: a decoupled pipeline that maintains real-time tracking while improving mapping-side rendering quality~\cite{huang2024photoslam}. 

%% file: 06_conclusions.tex
\section{Conclusion}
\label{sec:conclusion}

We studied online 3DGS mapping in a SLAM setting and found that several widely used \emph{offline} heuristics become brittle when optimization is constrained to a short per-keyframe budget. Building on the decoupled Photo-SLAM architecture~\cite{huang2024photoslam}, we proposed three lightweight, training-free mapping-side modifications: Transmittance-Preserving Densification to prevent opacity drift under cloning, Camera-aware Scale initialization to tie the initial Gaussian size to camera geometry rather than point density, and Error-guided Densification to explicitly allocate new primitives to regions with persistently high residual values.\del{All modifications operate in the mapping backend only and preserve the tracking frontend unchanged.} Finally, our results reinforce a methodological point: because SLAM pipelines are non-deterministic, reporting multi-run variance and using more demanding rendering protocols (e.g., full-trajectory evaluation rather than keyframe-only train-view rendering) is important for honest benchmarking in the rapidly growing 3DGS-SLAM literature. Future work could incorporate uncertainty-aware densification and explore principled evaluation protocols that combine novel-view rendering with trajectory reliability.


